\documentclass[conference]{IEEEtran}
\IEEEoverridecommandlockouts
\usepackage{cite}
\usepackage{amsmath,amssymb,amsfonts}
\usepackage{algorithmic}
\usepackage{graphicx}
\usepackage{textcomp}
\usepackage{amsmath}

\def\BibTeX{{\rm B\kern-.05em{\sc i\kern-.025em b}\kern-.08em
    T\kern-.1667em\lower.7ex\hbox{E}\kern-.125emX}}
\begin{document}
\title{Pseudorehearsal in actor-critic agents with neural network function approximation\\
}
\author{\IEEEauthorblockN{Vladimir Marochko}
\IEEEauthorblockA{
\textit{LLC Innodata}\\
Innopolis, Russia \\
vmarochko@innodata.ru}
\and
\IEEEauthorblockN{Leonard Johard}
\IEEEauthorblockA{
\textit{Innopolis University}\\
Innopolis, Russia \\
l.johard@innopolis.ru}
\and
\IEEEauthorblockN{Manuel Mazzara}
\IEEEauthorblockA{
\textit{Innopolis University}\\
Innopolis, Russia \\
m.mazzara@innopolis.ru}
\and
\IEEEauthorblockN{Luca Longo}
\IEEEauthorblockA{
\textit{Dublin Institute of Technology}\\
Dublin, Ireland \\
luca.longo@dit.ie}
}

\maketitle
\begin{abstract}
Catastrophic forgetting has a significant negative impact in reinforcement learning. The purpose of this study is to investigate how pseudorehearsal can change performance of an actor-critic agent with neural-network function approximation. We tested agent in a pole balancing task and compared different pseudorehearsal approaches. We have found that pseudorehearsal can assist learning and decrease forgetting.
\end{abstract}
\section{Introduction}
Reinforcement learning (RL) is a growing area of biologically inspired machine learning techniques. RL is used to train agents how to act in unknown environments with unknown optimal behavior. Agents know only the goals they have to reach. Training is based on numeric feedback to agent's actions. The common practice is to give a positive reward when the final goal is reached or a negative one when the final goal became impossible to reach.\\
When number of states recognizable by the agent is infinite or too large to keep in memory---function approximations are used for state processing. One of common approximations used in RL is an artificial neural network (ANN). Neural networks are vulnerable to the problem known as catastrophic forgetting (CF) which causes information losses in network during retraining. In RL ANN retraining occurs often---from once per step to once per episode, and so CF has a significant effect. Pseudorehearsal is one of the methods used to prevent CF, and we are going to use it to improve speed and quality of training. We conducted an experiment on a simulation of the pole balancing cart to prove that pseudorehearsal significantly improves performance of actor-critic agent.
\section{Background}
\subsection{Reinforcement learning}
Reinforcement learning is a framework of machine-learning-based applications for action selection, policy improvement and state evaluation. RL is a natural concept used by living creatures. First researches on RL in nature appeared about century ago. Edward Thorndike found that learning is based on the ability of the animals to figure out results from the consequences of their behavior \cite{thorndike1898review}. Later B.F.Skinner researched learning by reinforcement and punishment more deeply \cite{skinner1958reinforcement}.\\
According to R.Sutton RL as a computer science concept was born in 1979 at the University of Massachusetts. It was the result of analysis, extension and application of the ideas from the work of Klopf A. Harry  "Brain function and adaptive systems: a heterostatic theory" \cite{sutton1998reinforcement}. Growth in computational powers and techniques, fast decreasing of computers prices and flexibility of agents make RL a popular and promising area. All RL algorithms are based on a simple consequence: to get the observation of the current state of the environment, to apply some rule to choose the next action to reach the goal, to receive reward or punishment and to improve the rule. Observation is a representation of environment that agent can get and process. State is an observation at some moment of time. Observed environment has to be assumed as a Markov Decision Process (MDP). MDP is a mathematical framework for modeling decision-making problems. To denote MDP at environment Markov Property should be satisfied: each state $s_t$ at any timestep \textit{t} is conditionally independent of all previous states $s_{t-n}, \forall n \in N$  and actions $a_{t-n}$, while the next state reached from current after some action applied should be definable.\\
State-action mapping rules in RL agents are expressed in policy function $\pi(s_t, a)$. Policy denotes probability of choosing the action \textit{a} being in state $s_t$. Policy may vary from simple conditional rules to a complex self-sufficient functions. The value function $v(s_t)$ is a way the agent can predict some expected total reward that can be reached from state $s_t$. The action-value function $Q(s_t,a)$ denotes expected total reward reached from state$s_t$ if agent chooses action $a$ for the next step. The value function is used to create the model of environment.\\
If the MDP is not fully observable and can't be kept in memory---value function, policy, or both can be presented by function approximation. It approximates the agent input from the observed sensory input. Learning on approximated states may provide descent learning quality, while an insufficient approximation may lead to serious performance degradation \cite{melo2008analysis}. One of the widely used approximations is an ANN. The network takes the state observation as the input and the update rule computed by the algorithm as the target. It is trained to be both the approximation for the state and the policy or value function for this approximation.
\subsection{Actor-critic algorithms}
Actor-critic model-based algorithms are a class of advanced on-policy RL algorithms which can model environment and construct optimal policy for the resulting model. It was proved that in case of properly chosen learning parameters it can be asymptotically optimal---agent's total reward is close to the maximal one in case of a larger number of steps \cite{manfred1987estimation}, if the policy convergence is not too fast, while choosing these proper parameters can be difficult\cite{konda1999actor}.
Actor-critic agents use policy-based actor for action selection and train actor on evaluation by critic. Critic is a temporal-difference (TD) learning algorithm that models environment. Actor-critic approaches require minimal computation for action selections and achieve high performance \cite{sutton1998reinforcement}.
\subsubsection{Actor} Actor-critic algorithms use the actor to choose actions based on the current state. Actor is typically a policy gradient function. Policy gradient is a model-free algorithm which does not use any models and does not evaluate the states. It returns policy $\pi (s_t, a_t) $ for the state $s_t$, and updates this state to distribution mapping function by the rule of gradient accent: $\Delta \pi (s_t, a_t) \sim \nabla J_{\pi}$, where J is the total expected reward:
$J = E \{\sum_k( \gamma^k r_k )\}$, where $\gamma$ is the discounting factor which is a parameter for agent's farseeing. If $\gamma $ is close to 1, the trained agent can predict reward many steps before, if $\gamma $ is close to zero---the agent gives more attention to closest actions.\\
The agent using policy gradient as a gradient method is guaranteed to converge to some optimal policy, but policy gradient methods like any other gradient method can fall to local optimum.\\
Policy gradient actors can be presented by a neural network, in this case a choice function is applied to the ANN output. In this paper we used the softmax function \eqref{eq:1}\\
\begin{equation}
P(a=a_j|\pi (s, a) ) = \frac{e^{\frac{\pi(s,a_j)}{\tau}}}{\sum_{i=0}^{|a|}e^{\frac{\pi(s,a_i)}{\tau}} } \label{eq:1}
\end{equation}
where $\tau$ is the exploration parameter (here temperature): the higher the $\tau$, the more explorational the policy. Exploration-exploitation dilemma is a problem the agent solves to find globally optimal policy in descent time. Exploration is using already known information about environment. Exploitation is visiting previously unvisited states. The dilemma is in proper balancing that lets agent reach optimal policy quickly, but doesn't make it walk around some local optima.\\
Using a single neural network both for implementation of policy gradient algorithm and for function approximation simplifies the update rule, and resulting formula becomes: $\Delta \theta_{\pi} \approx \alpha \frac{\delta \rho}{\delta \theta_{\pi}}$.
The policy neural networks weights are denoted as $\theta_{\pi} $, the expected reward as $\rho$ and the learning rate as $\alpha $\cite{sutton2000policy}. 
\subsubsection{Critic}
Policy gradient methods have good convergence properties, but they do not store information about the environment, so their performance is limited. Critics estimates the policy which is currently being followed by the actor. They provide a critique which takes the form of a TD error: $\delta = R + \gamma V(s_{t+1}) - V(s_t) $. It drives all learning in both the actor and the critic. TD-error is usually counted on value-functions, then it is used for policy evaluation, extracted to the action-value function as: $\delta = R + \gamma Q(s_{t+1}, a_{t+1}) - Q(s_t, a_t) $, which is an update for SARSA-learning algorithm. SARSA agents evaluate both states and current policy.
Critic is a state-value function, it keeps information about environment and predicts action outcome. On the other hand now agent explores environment for a longer time \cite{beitelspacher2006policy}. After each action selection critic determines whether things have gone better or worse than expected.\\
We used neural networks to represent both actor and critic. In our case the number of possible actions from each state is small and constant, so critics outputs can be used directly for the learning of the actor as a target. After some simplifications from regular-gradient update rule we have following rules for the actor-critic: $\theta_{t+1} = \Gamma (\theta _t + \beta \delta _t \psi _{s_t,a_t}) $, where $\theta_t$ is the weight of policy gradient, $\beta $ is the policy learning rate, $\delta_{t}$ is the TD-error or the critic's estimation of last action, $\psi_{s_t,a_t}  $ is the approximation function, $\Gamma (x)$ is the projection function, which could be ignored assuming the iterations remain bounded\cite{bhatnagar2007incremental}.
As the single structure is used both for policy and approximation, the actor's update is simplified to \eqref{eq:2}
\begin{equation}
\Delta \theta = \beta \delta_t \label{eq:2}
\end{equation}
So the update of actor is made strictly by the critic's estimation of action. Policy learning rate $\beta$ should be significantly less than $\alpha$---action values learning rate, so agent learns the model and evaluate states much faster than it changes its policy. Consequently the actor acts independently while exploring the environment, and the critic's remarks help the actor but do not paralyze its initiative.\\
Such segregation of duties helps to provide fast decision making and high chance of exploration. The policy converges much slower than learning the model. This issue allows to act close to greedy policy in highly evaluated states and look for new ones until it is possible to make an improvement.
\subsubsection{Biological roots of actor-critic algorithm}
The functions of dopamine circuits in human and animal brain are normally explained as the sources of motivation that provide intrinsic reward for action learned as leading to reward even if the external reward was not provided. By this way dopamine neurons accelerate motivation and decision making \cite{TalanovVDMD15,VALLVERDU2016}. Lack of dopamine leads to problems with learning for long-term reward and is often seen as the root of addictions and procrastination, but learning still can occur successfully in case of an immediate reward. From this point of view dopamine neurons work as the critic in actor-critic algorithm\cite{johard2014connectionist}. That allows animals and people to learn to act for a reward which will be received much later than actions provided without effort of will, but with a pleasure caused by intrinsic reward produced by the critics. It is interesting to note that the update rules for dopamine critics are based not on actual rewards, but only on differences between the actual total reward and the predicted one \cite{berridge2007debate}, so, the actor-critic algorithm implements the real function of a brain structure in a similar way.
\subsection{Catastrophic forgetting}
\subsubsection{Causes of catastrophic forgetting}
Catastrophic forgetting (CF) is a common problem in neural networks that is connected with memory consolidation. The problem occurs when a neural network trained to execute some tasks faces changing conditions or learns to execute a new task. Old information might be erased in the process of learning new data. In RL this problem is common and leads to serious performance falls. Network's learning repeats regularly, relearning occurs very often, and target is usually not the same for the same input, so CF can seriously damage performance \cite{cahill2011catastrophic}.\\
The cause of CF lies in the mathematical nature of neural networks. In linear networks forgetting happens because neural networks base their prediction from input data on the vector orthogonality \cite{frean1999catastrophic}. The change of error in the first learned set depends on the dot product between sets. If the orthogonality is low, the error grows. In case of a non-linear ANN the situation is not so clear, but tends to be similar---the CF becomes significant when the information is very distributed and highly overlapping between sets \cite{moe2005catastophic}.
\subsubsection{Methods of avoiding catastrophic forgetting}
Most common simple methods to avoid of catastrophic forgetting can be grouped into two general approaches.
First approach includes different rehearsal, pseudorehearsal, and similar methods, which create additional patterns and make agent learn on them and on new examples together. After learning with such method there exist grounds to expect the performance to be as good as if the initial training occurred on both sets at the same time, not one after another. Rehearsal methods keep items from the previously learned sets in the rehearsal buffer. They fight CF well \cite{robins1995catastrophic}, but require additional memory. Pseudorehearsal (PR) is similar to rehearsal, but instead of real items from the old sets agent uses generated pseudoitems. More complex methods like transfer learning and dueling networks require creating of additional structures.\\
The second group comprises methods like context biasing and activation sharpening. They update learning rules for hidden layers to protect some part of meaningful information from changing. Activation sharpening is increasing the activation of some most active hidden units \cite{french1992semi}. Context biasing changes the activations based on Hamming distance between old and new activation vectors \cite{french1994dynamically}. The newer EWS method explored in \cite{kirkpatrick2017overcoming} is grounded on similar theoretical basis. Classic reducing overlap techniques don't really help to avoid CF, but reduce the time needed for relearning on the base set. 
\subsection{Pseudorehearsal}
PR is a simple and computationally efficient method for solving CF problem which is proven to be successful in unsupervised learning \cite{robins1995catastrophic}, supervised learning problems \cite{robins2004sequential}, \cite{moe2005catastophic} and sometimes in reinforcement learning as well \cite{marochko2017pseudorehearsal}, \cite{cahill2011catastrophic}, \cite{baddeley2008reinforcement}. It is interesting to note that the results of Baddeley suggest, that the widely studied ill conditioning might not be the main bottleneck of reinforcement learning while CF may be.\\
The PR is a two-step process: the first step is the construction of the set of pseudopatterns and the second is training the network using pseudopatterns. The optimal way of creating pseudopattern inputs is the subject of additional research. The simple one, proposed by Robins, is to assign randomly each element of input vector a score of 0 or 1. Feeding this pseudoinputs through the neural network and saving its outputs helps to save the internal state of the network. These models have been proven highly effective by \cite{robins1995catastrophic}, the argument of the authors was that, although the input is completely random, the activation distributions on deeper levels of the network will be representative of the previously learned input data. We can use pseudopatterns for the correction of learning weights in learning with respect to orthogonality between the learned example and pseudovectors; or we can use them as batch vectors for batch backpropagation algorithms. The first method was used by Frean and Robins in their work \cite{frean1999catastrophic}. Working on a similar research on Q-learning agents we improved those equations to make them more easy for implementation \eqref{eq:3}
\begin{equation} \Delta w_i = err_{b_i} \frac{1} {pr} \sum_{j=1}^{pr} \frac{ b_i (x_{ij}\cdot x_{ij}) - x_{ij} (x_{ij}\cdot b_i) } {(b_i \cdot b_i) (x_{ij}\cdot x_{ij}) - (b_i\cdot x_{ij}) (b_i \cdot x_{ij})} \label{eq:3}
\end{equation}
We also checked whether learning pseudopatterns on hidden layers activations improves performance, because neural networks in \cite{frean1999catastrophic} were linear, and our network is non-linear. In this project we will test both ideas, with PR on output only and PR on each layer, to find out if it gives any improvements. 
\subsection{Biological inspiration of pseudorehearsal}
The physiological part of PR approach is in the enforcing of knowledge consolidation based on random signals. This is similar and related to the processes which take place in the brain during REM-sleep, where memory consolidation happens. During that process brain does at the same time learn, providing significant memories to long-term memory, and unlearn---which makes us forget less important memories \cite{robins1999consolidation}. Hattori in his work had also shown that hippocampus structure provides avoidance of CF by a dual-network PR approach using pseudopatterns produced by neocortical networks  \cite{hattori2014biologically}.
\section{Experiment}
\subsection{Environment}
\begin{figure}[h!]
    \centering
    \includegraphics[width=0.3\textwidth]{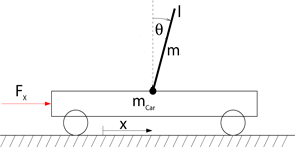}
    \caption{Double-pole cart}
    \label{fig:1}
\end{figure}

We apply PR algorithms to actor-critic agent executing the single-pole cart balancing problem, a well-known reinforcement learning task mentioned by Sutton \cite{sutton1998reinforcement} and extended further (Fig. \ref{fig:1}). The task is to balance a pole installed on a cart for as long as it is possible by pushing the cart left or right. If any pole falls or cart leaves the track the game is failed and the agent receives the reward R=-1. The output of this experiment is the number of steps the gent balanced the pole in an episode, the bigger this number the better. The dynamics is simulated by equations taken from Wieland \cite{wieland1991evolving} with the sign for angular acceleration changed---otherwise it was directed opposite to its physically expected direction.
\subsection{Agent}
The agent for learning on this task is an actor-critic agent using two feed-forward back-propagation neural networks, one is for actor and other is for critic. The learning rate of critic is $\alpha = 0.3$, the learning rate of actor is $\beta = 0.06$, the discounting factor for the intrinsic reward is $\gamma = 0.99$, the backpropagation algorithm for approaches with learning through batch-backpropagation is limited by time. We expect all PR approaches to improve agent's performance significantly in the long run.
\subsection{Observation}
The agent receives observation of the current cart's and poles' positions, velocities and accelerations. Observations are returned at discrete timesteps. The observation is represented as a real valued vector, where each $i^{th}$ observed parameter is written into one of two vector cells: $2*i^{th}$ if the parameter is positive or $(2*i+1)^{th}$ if parameter is negative. The second vector entity assigned to parameter is assigned to zero. After that, the linear parameters are divided by 20 and angular are divided by 60 for normalization. As the result is highly reliable on all parameters, and the actual reward is rare, and the environment is highly unstable, we have a low orthogonality of feature vectors from the agent's side. This environment with this observation suffers from a very high amount of CF. For this reason the environment is good for testing tools for elimination of CF.

For a more evident result representation we used plotting graphs of tendencies---vectors where $i^th$ element is mean of $i^th$ to $(i+100)^th$ elements of original result vector. Tendency graphs help to evaluate agent's average progress more accurately without the up and down jumps usual for the RL agents performance.\\
As the results are close to normally distributed, we applied the Student's T-test to the results to see if the results are statistically significant and therefore if we can make a strong statement based on research.
\section{Results}
Trying different force of push we have found that the agent and the PR have totally different behaviors with different force values. During the early training we found the most demonstrative results are reached at values 25 Newtons and 2.5 Newtons. A stronger push refers to high-risky actions, a weak push to low-risky ones. Low-risky environments have a larger and denser subspace of optimal states---states that have at least one possible action performing which leads to another optimal state. In a high-risky environment optimal states are placed much sparser.
\subsection{Learning agent with highly risky actions}
An environment with highly risky actions is a very complicated environment for the agent. Few actions will lead to success while nearly all the other will lead the agent to fail. Applying artificial agents to this type of tasks is very important, because the execution of such tasks by a human is highly stressful. Stress can seriously damage performance on complicated tasks \cite{bourne2003stress}. From this point of view even modest results in training will let to replace human agents with artificial ones in risky tasks with performance growth.
\subsubsection{Choosing better correction type for pseudorehearsal based learning}
\begin{figure}[h]
    \centering
    \includegraphics[width=0.5\textwidth]{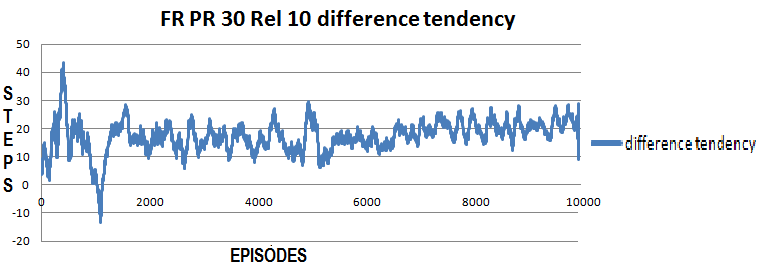}
    \caption{Tendency graph for the difference between the agent with PR correcting learning on outputs and one correcting learning on all layers}
    \label{fig:2}
\end{figure}

\begin{figure}[h]
    \centering
    \includegraphics[width=0.5\textwidth]{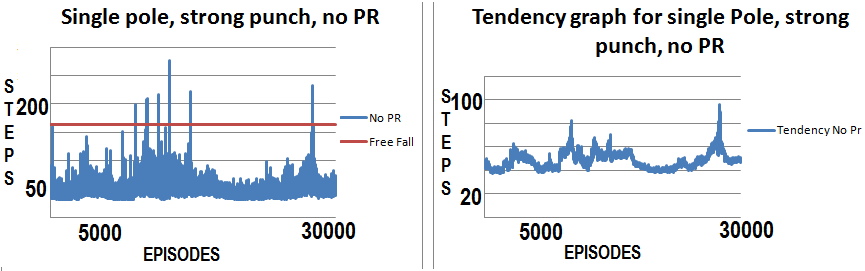}
    \caption{Even on initial graph it is perfectly visible where the learning 
occurs and where convergence to worse state takes place. The tendency graph just makes it more convenient for visual evaluation}
    \label{fig:3}
\end{figure}

\begin{figure}[h]
    \centering
    \includegraphics[width=0.5\textwidth]{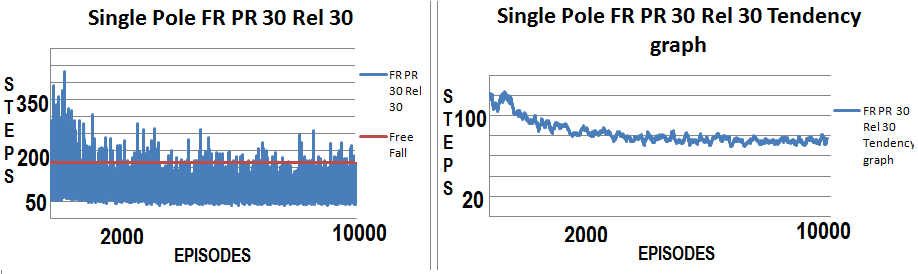}
    \caption{Typical steps-episode graph for FR PR with highly risky behaviour}
    \label{fig:4}
\end{figure}

\begin{figure}[h]
    \centering
    \includegraphics[width=0.5\textwidth]{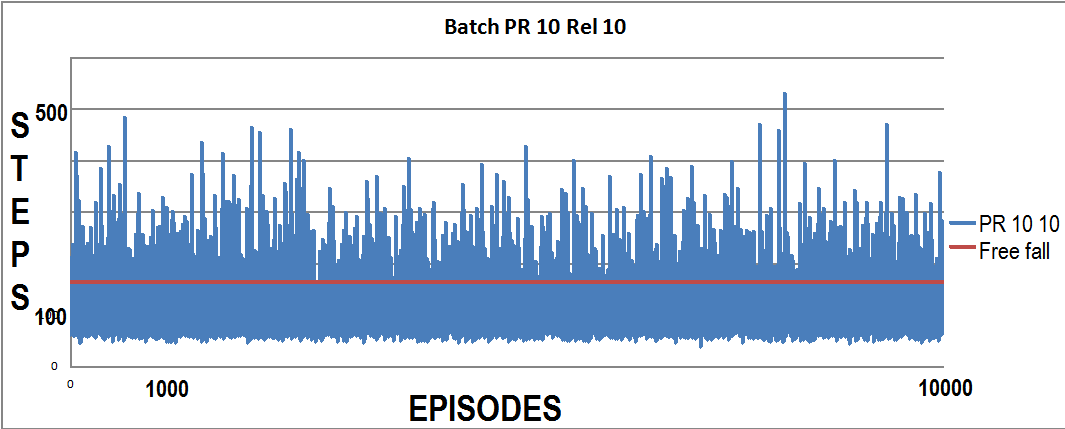}
    \caption{Batch-backpropagation based PR example}
    \label{fig:5}
\end{figure}

\begin{figure}[h]
    \centering
    \includegraphics[width=0.5\textwidth]{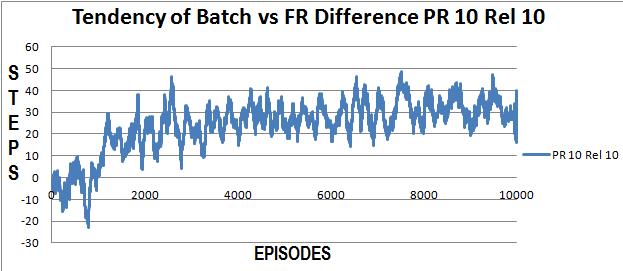}
    \caption{Comparing batch-backpropagation PR and FR PR by tendency graph.}
    \label{fig:6}
\end{figure}

\begin{figure}[h]
    \centering
    \includegraphics[width=0.5\textwidth]{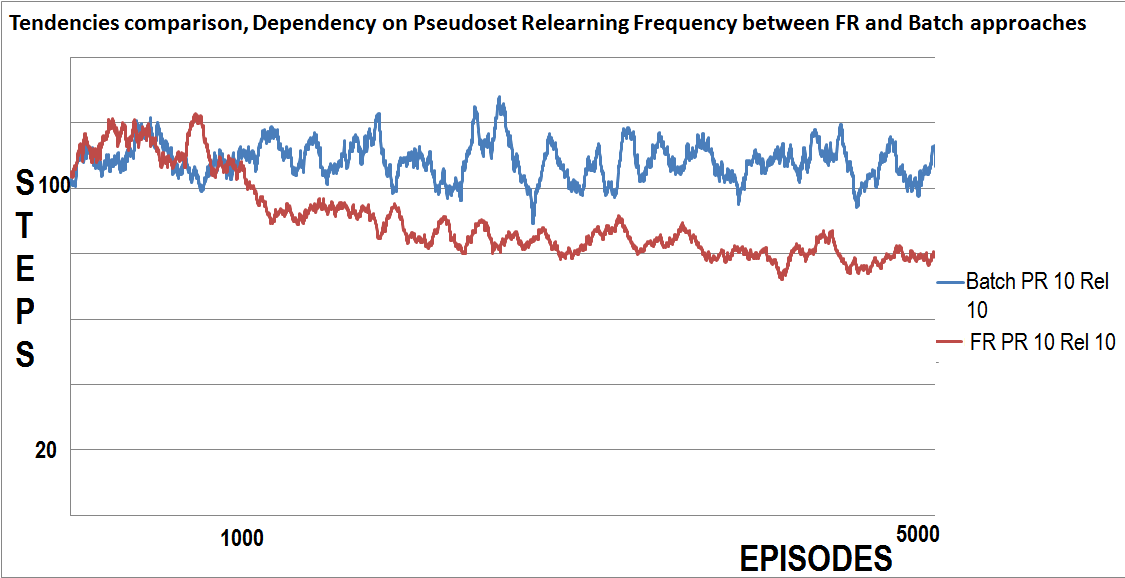}
    \caption{Comparing batch backpropagation PR and FR PR}
    \label{fig:7}
\end{figure}

The first question of interest is which kind of PR with learning rate correction (FR PR) in case of nonlinear network is better---with fixing learning rule only on output layer, or with fixing them on each layer w.r.t. pseudopattern's activations on those layers. For testing we took some samples for PR learning outputs and some with PR learning all activation. The comparison of samples with same parameters was provided by computing a difference vector, plotting its graph and plotting difference method tendency graph. For all parameters the visual evaluation has shown the same thing: FR PR applied to all layers have shown a much better result; all the tendencies graphs looked nearly the same. On Fig. \ref{fig:2} you may see the example for FR PR 30 Rel 10.
As we can see almost everywhere the tendency curve of difference is above zero. The performance of the FR PR agent with weight correction applied to all layers is higher. The results of significance test has shown that $t-stat \approx 43.77 >> $ t. The critical one-tail $\approx 1.645$ which means that this difference is statistically significant, therefore the using of FR PR applied on each layer of a non-linear neural network gives a much higher performance in all trials.\\
The actor-critic agent without PR starts with a poor policy as bad as the free fall behavior, performance falls quickly, then grows fast and later drops extremely. Then a similar behavioral template repeats again with even lower results. Almost all runs have a lower performance then if the pole were just falling down without any involvement. The convergence occurs on a very low local optimum (Fig.\ref{fig:3}). We suppose that CF has such effect because of a high number of negative rewards. Agent mistakenly evaluates possibly optimal states similar to risky ones like risky too. It erases the previously learned overlapping weights.
\\
In tests the performance of all agents was still poor, but better than in the no-RP case, and there were visible differences of behavior connected with types of PR. Sizes of pseudosets and reinitialization frequency show almost no effect on the behavior.\\
FR PR shows a strong improvement of performance: Fig. \ref{fig:4}
After the agent starts it quickly learns how to reach a fairly high performance two to three times better then free fall, but then agent starts to diverge slowly. Finally it diverges to a policy worse than the initially reached, but still better than the agent without PR has. After reaching that level no serious changes happen. This result shows that the FR PR helps the agent to widen the explorations and go further even in cases of meeting highly negative rewards. For the first time it protects weights from being immediately erased by an avalanche of negative rewards. Because of this protection the agent may continue exploration in these highly negative states for more time than the agent without PR.\\
The batch-backpropagation PR approach has shown a good result in avoiding CF. The overall performance is sufficiently high and doesn't suffer from slow monotonic degradation like in examples with FR PR (Fig.\ref{fig:5}).
Visual comparison of different PR approaches of PR with the same size of pseudoset and relearning frequency shows that batch-backpropagation provides a higher efficiency and a lower stability (Fig. \ref{fig:6}).
While the batch PR agent has a mean value higher by more than $25\%$ than the correcting weights agent: 107.6 vs 83.6, it has an about three times higher variance: 2900 vs 1080. That means that the agent with batch PR type has a far more aggressive and risky policy. This tendency to visit risky states seems to be caused by deeper and stronger relearning of previous networks internal state, so after CF occurrences the agent quickly returns to the previous high performance.\\
Comparing the tendency graphs, one can see that while initially the tendencies of weight-correcting approach were at least not worse than batch one, after some number of steps the weight-correcting starts to diverge to a less risky suboptimal state, while the batch PR is still at the top (Fig.\ref{fig:7}).
The result of the experiment shows that the agent oscillates around some quickly found local optimum. The Results of these oscillations are much higher than the properly results of FR PR and no PR approaches. The significance test has approved that difference between 
approaches has $\sim$ 38 as t-statistics result---a significance so high that it was obvious even before the test. The computational cost of the batch-backpropagation algorithm implementation used in the test is one-two orders of magnitude higher than the computational time of vector multiplications used for learning rate correcting.
\subsection{Learning agent with low risky actions}
\begin{figure}[h]
    \centering
    \includegraphics[width=0.5\textwidth]{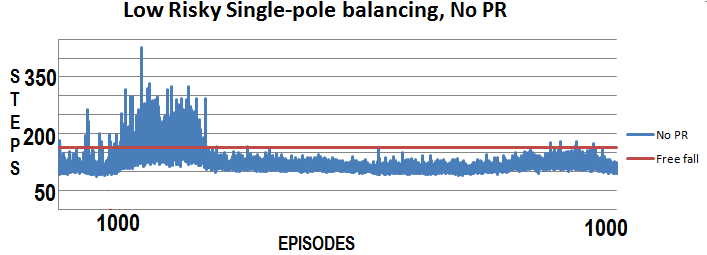}
    \caption{Agents performance without PR and with low risky actions}
    \label{fig:8}
\end{figure}

\begin{figure}[h]
    \centering
    \includegraphics[width=0.5\textwidth]{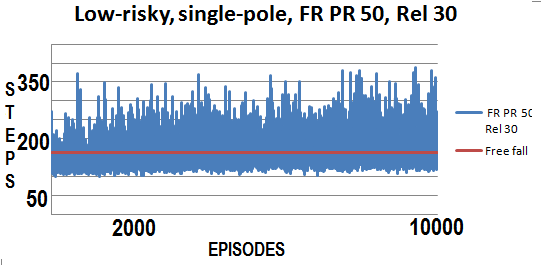}
    \caption{Agents performance with learning-rate correction PR}
    \label{fig:9}
\end{figure}

\begin{figure}[h]
    \centering
    \includegraphics[width=0.5\textwidth]{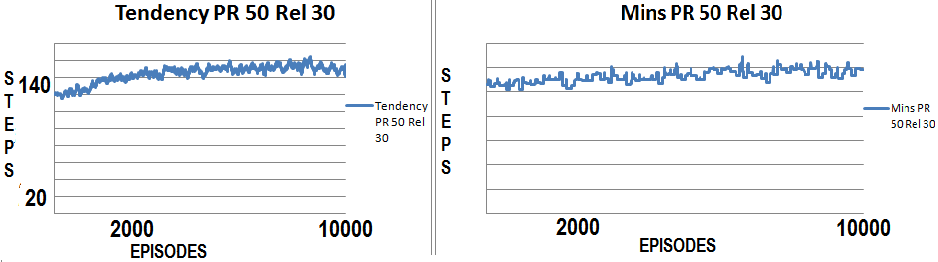}
    \caption{Tendencies of mean and minimal values for agent with FR PR}
    \label{fig:10}
\end{figure}

\begin{figure}[h]
    \centering
    \includegraphics[width=0.5\textwidth]{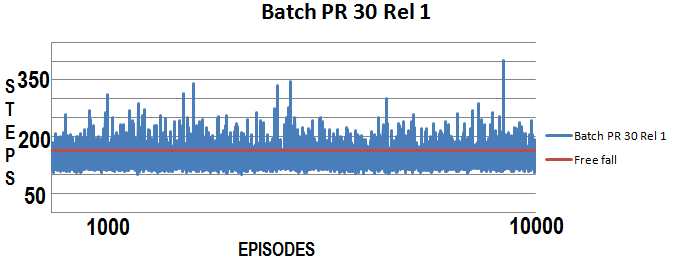}
    \caption{Performance graph for batch-backpropagation approach}
    \label{fig:11}
\end{figure}

In a less risky case there exists a large subspace within the state space, where almost all actions will not lead to the falling of the pole. As the agent moves further from the center of this subspace, the higher is chance to perform an action which will drop the pole. When the size of safe state or states is increased there is a lower risk that the values learned as safe will be overwritten because of an overlapping with a very similar state. In this type of problem it's harder to lose all the knowledge collected because of CF. On the other hand in current problem the time period between reaching failing state and actual failing of the task is larger.\\
Acting in a low-risky environment is simpler and the actor-critic agent without PR has reached high performance fast and was showing the same high performance for about two thousand steps. The agent also increased the avoidance of dangerous states, marked by the increased lower boundary on the graph. This agent is better in recognizing states with lowest expected reward and avoiding them. After that sequence of good policy the agent's performance quickly diverges and can't return to the same high result again (Fig.\ref{fig:8}). 
To avoid this the performance drop caused by CF PR is used. The FR PR in this type of problems shows a very interesting picture: not only it coherently grows, but its lower boundary goes up as well (Fig.\ref{fig:9}).
We plotted tendencies graph and graph of smoothed minimums denoted by the following rule: $i^{th}$ element of minimums vector is $min(i^{th}, i+100^{th})$ from original sample. Both graphs grow coherently, and are expected to converge to some optimal policy with high performance (Fig.\ref{fig:10}).
The batch PR approach worked far worse than FR PR and worse than in case of an environment with highly risky actions. The performance was a bit better than the free fall, and significantly better than the agent without PR had, but much lower than in case of FR PR. The value of mean is about 140 vs 163 which is 1.15 times lower. As well the batch PR agent seems to reach its optimal policy and neither learning nor relearning occurs any more---it keeps oscillating around same value for most part of experiment---except initial learning at the very beginning of learning series (Fig.\ref{fig:11}). The T-test proved this difference to be significant.
Applying the learning rate correction has smoothed learning curve and improved learning with a significance (T-Stat) about 4.5. It's interesting to note, that unlike all previous approaches it did not improve the mean of performance much. Improvement was neglectable---from 56.39 to 56.99.  This PR approach decreased variance from 101.7 to 69.9---about 1.45 times. As the picture shows there are less high spikes, but the overall performance is converged to some local optimum and does not fall lower.
\section{Conclusion and future work}
We proved that the PR approaches strongly improve the learning of an actor-critic agent with softmax action-selection function. Different PR approaches have different ways and different performance boosts, but all of them were statistically significant and none of them led to a worse performance. The possible cause is that the actor-critic agent has a higher possibility of exploration of promising states.\\
We used a nonlinear neural network with hidden layer and proved that activations returned by neurons in the hidden layer for pseudoinputs improved performance significantly compared to keeping activations on output layer only.  The statistical significance test has shown 43.77 for T-stats which is a highly significant result.\\
We found that the effects of different PR approaches vary in different situations which depends on the density of distribution of optimal subspaces in multidimensional state space. This dependency was found during the early training when choosing optimal force of push for the algorithm to make graphs maximally evident. When talking about environments with high risky and low risky actions, the possible cost of mistake is denoted in the very terms. The experiment has shown that the batch-backpropagation algorithms provide a better performance that does not degrade in high-risky environments. In a low-risky environment FR PR trains faster and the performance continues growing after initial training. These results can help in choosing the PR approach for a concrete task. As this topic is not thoroughly explored, it might lead to many new interesting experiments and discoveries in reinforcement learning agents.\\
Some issues remain open for research. First of all, the exploration of different RL algorithms in high-risky and low-risky environments. It would help to find important properties of environment and to choose proper algorithms and parameters to task execution. Secondly, it is the application of PR to different reinforcement and supervised learning approaches, finding dependencies, similarities and differences between them and creating a mathematical foundation which can help to choose an optimal approach to current task. Potential application scenarios can be seen in innovative technologies, such as smart houses \cite{Nalin2016} and smart automotive systems \cite{Gmehlich13}. As it was found that in actor-critic algorithms PR parameters give no any significant difference in agents performance, while in other algorithms they might have a much higher meaning, it is necessary to find which parameters of agent or environment make PR parameters more or less meaningful, and find out a way to easily predict those parameters and reduce their significance.
\bibliographystyle{IEEEtran}
\bibliography{IEEEabrv,mainbib}
\end{document}